\documentstyle[aclap]{article}

\include{epsf} 

\title{Developing a hybrid NP parser}

\author{Atro Voutilainen \\
Department of General Linguistics \\
P.O. Box 4 \\ 
FIN-00014 University of Helsinki \\
Finland \\{\tt avoutila@ling.helsinki.fi}\And
Llu\'{\i}s Padr\'o \\
Dept. Llenguatges i Sistemes Inform\`atics \\
Universitat Polit\`ecnica de Catalunya \\
C/ Gran Capit\`a s/n.  08034 Barcelona \\
Catalonia \\{\tt padro@lsi.upc.es}}

\begin{document}
\bibliographystyle{fullname}
\maketitle

\begin{abstract}

  We describe the use of energy function optimization in very shallow
  syntactic parsing. The approach can use linguistic rules and
  corpus-based statistics, so the strengths of both linguistic and
  statistical approaches to NLP can be combined in a single framework.
  The rules are contextual constraints for resolving syntactic
  ambiguities expressed as alternative tags, and the statistical
  language model consists of corpus-based n-grams of syntactic tags.
  The success of the hybrid syntactic disambiguator is evaluated
  against a held-out benchmark corpus. Also the contributions of the
  linguistic and statistical language models to the hybrid model are
  estimated.

\end{abstract}

\section{Introduction}

The language models used by natural language analyzers are
traditionally based on two approaches. In the linguistic approach, the
model is based on hand-crafted rules derived from the linguist's
innate and/or corpus-based knowledge about the object language. In the
data-driven approach, the model is automatically generated from
annotated text corpora, and the model can be represented e.g.\ as
ngrams \cite{Garside87}, local rules \cite{Hindle89} or neural nets
\cite{Schmid94}.

\medskip

Most hybrid approaches combine statistical information with
automatically extracted rule-based information
\cite{Brill95,Daelemans96}.  Relatively little attention has been paid
to models where the statistical approach is combined with a truly
linguistic model (i.e.\ one generated by a linguist).  This paper
reports one such approach: syntactic rules written by a linguist are
combined with statistical information using the relaxation labelling
algorithm.

\medskip

Our application is very shallow parsing: identification of verbs,
premodifiers, nominal and adverbial heads, and certain kinds of
postmodifiers. We call this parser a noun phrase parser.

The input is English text morphologically tagged with a rule-based
tagger called EngCG \cite{Voutilainen92,Karlsson95}. Syntactic
word-tags are added as alternatives (e.g. each adjective gets a
premodifier tag, postmodifier tag and a nominal head tag as
alternatives). The system should remove contextually illegitimate tags
and leave intact each word's most appropriate tag. In other words, the
syntactic language model is applied by a disambiguator.

The parser has a {\em recall} of 100\% if all words retain the correct
morphological and syntactic reading; the system's {\em precision} is
100\% if the output contains no illegitimate morphological or
syntactic readings. In practice, some correct analyses are discarded,
and some ambiguities remain unresolved.

\medskip

The system can use linguistic rules and corpus-based statistics.
Notable about the system is that minimal human effort was needed for
creating its language models (the linguistic consisting of syntactic
disambiguation rules based on the Constraint Grammar framework
\cite{Karlsson90,Karlsson95}; the corpus-based consisting of bigrams
and trigrams):

\begin{itemize}

\item Only one day was spent on writing the 107 syntactic
  disambiguation rules used by the linguistic parser.

\item No human annotors were needed for annotating the training corpus
  (218,000 words of journalese) used by the data-driven learning
  modules of this system: the training corpus was annotated by (i)
  tagging it with the EngCG morphological tagger, (ii) making the
  tagged text syntactially ambiguous by adding the alternative
  syntactic tags to the words, and (iii) resolving most of these
  syntactic ambiguities by applying the parser with the 107
  disambiguation rules.

\end{itemize}

\medskip

\noindent The system was tested against a fresh sample of five texts (6,500
words). The system's recall and precision was measured by comparing
its output to a manually disambiguated version of the text. To
increase the objectivity of the evaluation, system outputs and the
benchmark corpus are made publicly accessible (see Section 6).

Also the relative contributions of the linguistic and statistical
components are evaluated. The linguistic rules seldom discard the
correct tag, i.e.\ they have a very high recall, but their problem is
remaining ambiguity. The problems of the statistical components are
the opposite: their recall is considerably lower, but more (if not
all) ambiguities are resolved. When these components are used in a
balanced way, the system's overall recall is \(97.2\%\) -- that is,
\(97.2\%\) of all words get the correct analysis -- and its precision
is \(96.1\%\) -- that is, of the readings returned by the system,
\(96.1\%\) are correct.

The system architecture is presented in Figure~1.

\medskip\noindent\hfil\epsfbox{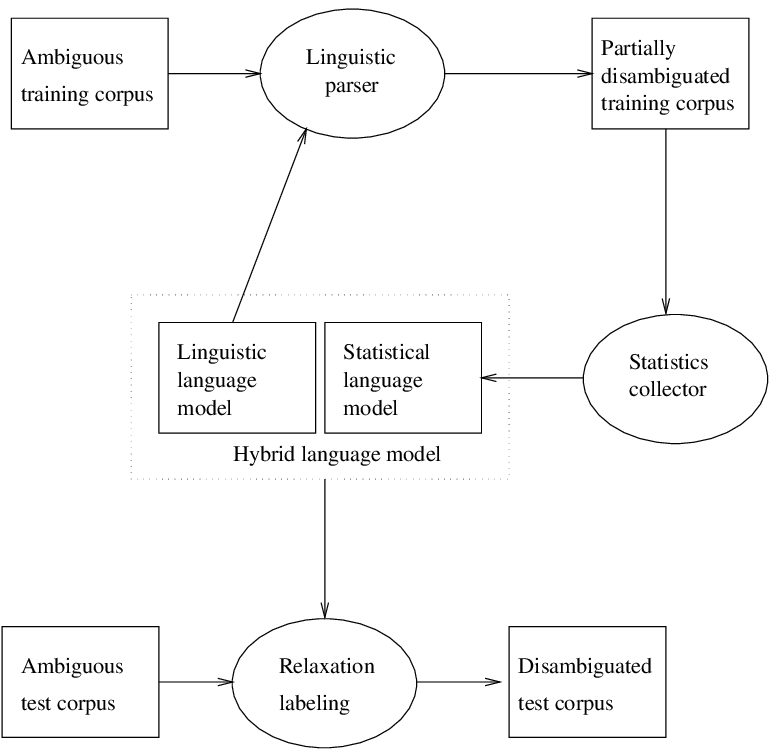}\hfil

\hfil \small{Figure~1: Parser architecture.} \hfil
\normalsize

\medskip

The structure of the paper is the following. First, we describe our
general framework, the relaxation labelling algorithm. Then we proceed
to the application by outlining the grammatical representation used in
our shallow syntax. After this, the disambiguation rules and their
development are described. Next in turn is a description of how the
data-driven language model was generated. The evaluation of the system
is then presented: first the preparation of the benchmark corpus is
described, then the results of the tests are given. The paper ends
with some concluding remarks.

\section{The Relaxation Labelling Algorithm}
\label{section-relax}

Since we are dealing with a set of constraints and want to find a
solution which optimally satisfies them all, we can use a standard
Constraint Satisfaction algorithm to solve that problem.

Constraint Satisfaction Problems are naturally modelled as Consistent
Labeling Problems \cite{Larrosa95}. An algorithm that solves CLPs is
Relaxation Labelling.

It has been applied to part-of-speech tagging \cite{Padro96} showing
that it can yield as good results as a HMM tagger when using the same
information. In addition, it can deal with any kind of constraints,
thus the model can be improved by adding any other
constraints available, either statistics, hand-written or automatically
extracted \cite{Marquez95,Samuelson96}.

Relaxation labelling is a generic name for a family of iterative
algorithms which perform function optimization, based on local
information.  See \cite{Torras89} for a summary.

Given a set of variables, a set of possible labels for each variable,
and a set of compatibility constraints between those labels, the
algorithm finds a combination of weights for the labels that maximizes
``global consistency'' (see below). \medskip
 
  Let \(V=\{v_1,v_2,\ldots,v_n\}\) be a set of variables.

  Let \(t_i=\{t_1^i,t_2^i,\ldots,t_{m_i}^i\}\) be the set of possible
  labels for variable \(v_i\).

  Let \(CS\) be a set of constraints between the labels of the variables.
  Each constraint \(C\in CS\) states a ``compatibility value'' \(C_r\) for 
  a combination of pairs variable--label. 
  Any number of variables may be involved in a constraint. 
  
The aim of the algorithm is to find a weighted labelling\footnote{A
    weighted labelling is a weight assignment for each label of each
    variable such that the weights for the labels of the same variable
    add up to one.}
such that ``global consistency'' is maximized. Maximizing ``global
consistency'' is defined as maximizing
\(\sum_j p^i_j \times S_{ij} \) ,  \(\forall v_i\),
where \(p^i_j\) is the weight for label \(j\) in variable \(v_i\) and
\(S_{ij}\) the support received by the same combination. The support
for the pair variable--label expresses {\em how compatible} that
pair is with the labels of neighbouring variables, according to
the constraint set. 

The support is defined as the sum of the influence of every constraint
on a label.
$$
S_{ij}={\displaystyle\sum_{r \in R_{ij}}{\mbox{Inf}(r)}} 
$$
where: \\
\(R_{ij}\) is the set of constraints on label \(j\) for variable 
\(i\), i.e. the constraints formed by any combination of
variable--label pairs that includes the pair \((v_i,t^i_j)\).\\
\(\mbox{Inf}(r) = C_r \times p^{r_1}_{k_1}(m) \times \ldots
\times p^{r_d}_{k_d}(m)\), is the product of
the current weights\footnote{\(p^{r}_{k}(m)\) is the weight assigned to label
\(k\) for variable \(r\) at time \(m\).} for the labels appearing in the constraint except
\((v_i,t^i_j)\) (representing {\em how
applicable} the constraint is in the current context) 
multiplied by \(C_r\) which is the constraint compatibility value
(stating {\em how compatible} the pair is with the context). 
\medskip

  Briefly, what the algorithm does is:
\begin{enumerate}
\item Start with a random weight assignment.
\item Compute the support value for each label of each variable.
         (How compatible it is with the current weights for the labels of the other variables.)
\item Increase the weights of the labels more compatible with the
         context (support greater than \(0\)) and decrease those of
         the less compatible labels (support less than
         \(0\))\footnote{Negative values for support indicate  {\em incompatibility}.},
         using the updating function:
$$
p^i_j(m+1) = \frac{{\displaystyle p^i_j(m)
    \times(1+S_{ij})}}{{\displaystyle
    \sum_{k=1}^{k_i}{p^i_k(m)\times(1+S_{ik})}}}
$$
$$ \;\;\;\;\;\;\;\;\;\; {\rm where} \;\; -1\le S_{ij}\le +1 $$
\item If a stopping/convergence criterion\footnote{The usual
         criterion is to stop when there are no more changes, although
         more sophisticated heuristic procedures are also used to stop
         relaxation processes \cite{Eklundh78,Richards81}.} is satisfied,
         stop, otherwise go to to step 2.
\end{enumerate}

\section{Grammatical representation}

The input of our parser is morphologically analyzed and disambiguated
text enriched with alternative syntactic tags, e.g.

\begin{verbatim}
"<others>"
   "other" PRON NOM PL @>N @NH
"<moved>"
   "move" <SV> <SVO> V PAST VFIN @V
"<away>"
   "away" ADV ADVL @>A @AH
"<from>"
   "from" PREP @DUMMY
"<traditional>"
   "traditional" A ABS @>N @N< @NH
"<jazz>"
   "jazz" <-Indef> N NOM SG @>N @NH
"<practice>"
   "practice" N NOM SG @>N @NH
   "practice" <SVO> V PRES -SG3 VFIN @V
\end{verbatim}

Every indented line represents a morphological analysis; the sample
shows that some morphological ambiguities are not resolved by the
rule-based morphological disambiguator, known as the EngCG tagger
\cite{Voutilainen92,Karlsson95}.

Our syntactic tags start with the "@" sign. A word is syntactically
ambiguous if it has more than one syntactic tags (e.g. {\it practice}
above has three alternative syntactic tags).  Syntactic tags are added
to the morphological analysis with a simple lookup module.  The
syntactic parser's main task is disambiguating (rather than adding new
information to the input sentence): contextually illegitimate
alternatives should be discarded, while legitimate tags should be
retained (note that also morphological ambiguities may be resolved as
a side effect).

\medskip

Next we describe the syntactic tags:

\begin{itemize}

\item @$>$N represents premodifiers and determiners.

\item @N$<$ represents a restricted range of postmodifiers and the
determiner "enough" following its nominal head.

\item @NH represents nominal heads (nouns, adjectives, pronouns,
  numerals, ING-forms and non-finite ED-forms).

\item @$>$A represents those adverbs that premodify (intensify) adjectives
(including adjectival ING-forms and non-finite ED-forms), adverbs and
various kinds of quantifiers (certain determiners, pronouns and
numerals).

\item @AH represents adverbs that function as head of an adverbial phrase.

\item @A$<$ represents the postmodifying adverb "enough".

\item @V represents verbs and auxiliaries (incl.\ the infinitive marker
"to").

\item @$>$CC represents words introducing a coordination ("either",
"neither", "both").

\item @CC represents coordinating conjunctions.

\item @CS represents subordinating conjunctions.

\item @DUMMY represents all prepositions, i.e.\ the parser does not
  address the attachment of prepositional phrases.

\end{itemize}

\section{Syntactic rules}
\label{section-CG}
\subsection{Rule formalism}

The rules follow the Constraint Grammar formalism, and they were
applied using the recent parser-compiler CG-2 \cite{Tapanainen96}. The
parser reads a sentence at a time and discards those
ambiguity-forming readings that are disallowed by a constraint.
\medskip

Next we describe some basic features of the rule formalism. The rule

\begin{verbatim}
REMOVE (@>N)
 (*1C <<< OR (@V) OR (@CS) BARRIER (@NH));
\end{verbatim}

\noindent removes the premodifier tag @$>$N from an ambiguous reading if
somewhere to the right (*1) there is an unambiguous (C) occurrence of
a member of the set $<<<$ (sentence boundary symbols) or the verb tag
@V or the subordinating conjunction tag @CS, and there are no
intervening tags for nominal heads (@NH).  \medskip

This is a partial rule about coordination:

\begin{verbatim}
REMOVE (@>N)
 (NOT 0 (DET) OR (NUM) OR (A))
 (1C (CC))
 (2C (DET)) ;
\end{verbatim}

\noindent It removes the premodifier tag if all three context-conditions are satisfied:

\begin{itemize}

\item the word to be disambiguated (0) is not a determiner, numeral or adjective,

\item the first word to the right (1) is an unambiguous coordinating
  conjunction, and

\item the second word to the right is an unambiguous determiner.

\end{itemize}

The rules can refer to words and tags directly or by means of
predefined sets. They can refer not only to any fixed context
positions; also reference to contextual patterns is possible.  The
rules never discard a last reading, so every word retains at least one
analysis. On the other hand, an ambiguity remains unresolved if there
are no rules for that particular type of ambiguity.

\subsection{Grammar development}

A day was spent on writing 107 constraints; about 15,000 words of the
parser's output were proofread during the process. The routine was the
following:

\begin{enumerate}

\item The current grammar (containing e.g.\ 2 rules) is applied to the
ambiguous input in a `trace' mode in which the parser also indicates,
which rule discarded which analysis,

\item The grammarian observes remaining ambiguities and proposes new rules
for disambiguating them, and

\item He also tries to identify misanalyses (cases where the correct
  tag is discarded) and, using the trace information, corrects the
  faulty rule

\end{enumerate}

This routine is useful if the development time is very restricted, and
only the most common ambiguity types have to be resolved with
reasonable success.  However, if the grammar should be of a very high
quality (extremely few mispredictions, high degree of ambiguity
resolution), a large test corpus, formally similar to the input except
for the manually added extra information about the correct analysis,
should be used. This kind of test corpus would enable the automatic
identification of mispredictions as well as counting of various
performance statistics for the rules. However, manually disambiguating
a test corpus of a few hundred thousand words would probably require a
human effort of at least a month.

\subsection{Sample output}

The following is genuine output of the linguistic (CG-2) parser using
the 107 syntactic disambiguation rules. The traces starting with "S:"
indicate the line on which the applied rule is in the grammar file.
One syntactic (and morphological) ambiguity remains unresolved: {\it
  until} remains ambiguous due to preposition and subordinating
conjunction readings.

\begin{verbatim}
"<aachen>" S:46
   "aachen" <*> <Proper> N NOM SG @NH
"<remained>"
   "remain" <SVC/N> <SVC/A> V PAST VFIN @V
"<a>"
   "a" <Indef> DET CENTRAL ART SG @>N
"<free>" S:316, 49
   "free" A ABS @>N
"<imperial>" S:49, 57
   "imperial" A ABS @>N
"<city>" S:46
   "city" N NOM SG @NH
"<until>"
   "until" PREP @DUMMY
   "until" <**CLB> CS @CS
"<occupied>" S:116, 345, 46
   "occupy" <SVO> PCP2 @V
"<by>"
   "by" PREP @DUMMY
"<france>" S:46
   "france" <*> <Proper> N NOM SG @NH
"<in>"
   "in" PREP @DUMMY
"<1794>" S:121, 49
   "1794" <1900> NUM CARD @NH
"<$.>"
\end{verbatim}

\section{Hybrid language model}

  To solve shallow parsing with the relaxation labelling algorithm
  we model each word in the sentence as a variable, and each of its
  possible readings as a label for that variable. We start with a
  uniform weight distribution.

   We will use the algorithm to select the right syntactic tag for
   every word. Each iteration will increase
   the weight for the tag which is currently most compatible with the
   context and decrease the weights for the others. 

    Since constraints are used to decide {\em how compatible} a tag is with
    its context, they have to assess the compatibility of a combination of
  readings. We adapt CG constraints described above.
\medskip

The {\bf REMOVE} constraints ex\-press to\-tal in\-com\-pa\-ti\-bi\-li\-ty\footnote{We
  model compatibility values using mutual information \cite{Cover91},
  which enables us to use negative numbers to state {\em
    incompatibility}. See \cite{Padro96} for a performance comparison
  between M.I.\ and other measures when applying relaxation labelling to
  NLP.}  and {\bf SE\-LECT} con\-straints ex\-press to\-tal
com\-pa\-ti\-bi\-li\-ty (actual\-ly, they ex\-press in\-com\-pa\-ti\-bi\-li\-ty of all
ot\-her pos\-si\-bi\-li\-ties).

   The compatibility value for these should be at least as strong as
   the strongest value for
   a statistically obtained constraint (see below). This
   produces a value of about $\pm 10$. 

   But because we want the linguistic part of the model to be more
   important than the statistical part and because a given label will
   receive the influence of about two bigrams and three
   trigrams\footnote{The algorithm tends to select one label per
     variable, so there is always a bi/trigram which is applied more
     significantly than the others.}, a single linguistic constraint
   might have to override five statistical constraints. So we will
   make the compatibility values six times stronger, that is, $\pm
   60$. \medskip

   Since in our implementation of the CG parser \cite{Tapanainen96}
   constraints tend to be applied in a certain order -- e.g.\ {\bf
     SELECT} constraints are usually applied {\em before} {\bf REMOVE}
   constraints -- we adjust the compatibility values to get a similar
   effect: if the value for {\bf SELECT} constraints is $+60$, the
   value for {\bf REMOVE} constraints will be lower in absolute value,
   (i.e.\ $-50$). With this we ensure that two contradictory
   constraints (if there are any) do not cancel each other.  The {\bf
     SELECT} constraint will win, as if it had been applied before.
   \medskip

This enables using any Constraint Grammar with this algorithm although
we are applying it more flexibly: we do not decide whether a
constraint is applied or not. It is always applied with an influence
(perhaps zero) that depends on the weights of the labels.

If the algorithm should apply the constraints in a more strict way, we
can introduce an influence threshold under which a constraint does not
have enough influence, i.e.\ is not applied.  \medskip

   We can add more information to our model in the form of
   statistically derived constraints. Here we use bigrams and
   trigrams as constraints.

   The 218,000-word corpus of
   journalese from which these constraints were extracted was
   analysed using the following modules:

\begin{itemize}

\item EngCG morphological tagger
\item Module for introducing syntactic ambiguities
\item The NP disambiguator using the 107 rules written in a day

\end{itemize}

No human effort was spent on creating this
   training corpus.
   The training corpus is partly ambiguous, so the
   bi/trigram information acquired will be slightly noisy, but
   accurate enough to provide an {\em almost} supervised statistical model.
 
   For instance, the following constraints have been statistically
   extracted from bi/trigram occurrences in the training corpus. 

\begin{verbatim}
-0.415371 (@V)
        (1 (@>N));

4.28089 (@>A)
        (-1 (@>A))
        (1 (@AH));
\end{verbatim}

   The compatibility value 
   is the mutual information, computed from the probabilities
   estimated from a training corpus. We do not need to assign the
   compatibility values here, since we can estimate them from the corpus.
\medskip

   The compatibility values assigned to the hand-written constraints
   express the strength of these constraints compared to the
   statistical ones. Modifing those values means changing the relative
   weights of the linguistic and statistical parts of the model.

\section{Preparation of the benchmark corpus}

For evaluating the systems, five roughly equal-sized benchmark corpora
not used in the development of our parsers and taggers were prepared.
The texts, totaling 6,500 words, were copied from the Gutenberg e-text
archive, and they represent present-day American English. One text is
from an article about AIDS; another concerns brainwashing techniques;
the third describes guerilla warfare tactics; the fourth addresses the
assassination of J.~F.~Kennedy; the last is an extract from a speech
by Noam Chomsky.

The texts were first analysed by a recent version of the morphological
analyser and rule-based disambiguator EngCG, then the syntactic
ambiguities were added with a simple lookup module. The ambiguous text
was then manually disambiguated. The disambiguated texts were also
proofread afterwards.  Usually, this practice resulted in one analysis
per word. However, there were two types of exception:

\begin{enumerate}

\item The input did not contain the desired alternative (due to a
  morphological disambiguation error). In these cases, no reading was
  marked as correct. Two such words were found in the corpora; they
  detract from the performance figures.

\item The input contained more than one analyses all of which seemed
  equally legitimate, even when semantic and textual criteria were
  consulted. In these cases, all the equal alternatives were marked
  as correct. The benchmark corpus contains 18 words (mainly ING-forms and
  nonfinite ED-forms) with two correct syntactic analyses.

\end{enumerate}

The number of multiple analyses could probably be made even smaller by
specifying the grammatical representation (usage principles of the
syntactic tags) in more detail, in particular incorporating some
analysis conventions for certain apparent borderline cases (for a
discussion of specifying a parser's linguistic task, see
\cite{Voutilainen95}).

To improve the objectivity of the evaluation, the benchmark corpus (as
well as parser outputs) have been made available from the following
URLs:\\ http:\-//www.\-ling.\-helsinki.\.fi/\~{
  }avoutila/\-anlp97.html\\ http:\-//www--lsi.\-upc.\-es/\~{
  }lluisp/\-anlp97.html

\section{Experiments and results}

We tested linguistic, statistical and hybrid language models, using
the CG-2 parser \cite{Tapanainen96} and the relaxation labelling
algorithm described in Section~\ref{section-relax}.

\medskip

The statistical models were obtained from a training corpus of 218,000
words of journalese, syntactically annotated using the linguistic
parser (see above).  

Although the linguistic CG-2 parser does not disambiguate completely,
it seems to have an almost perfect recall (cf.\ Table 1 below), and
the noise introduced by the remaining ambiguity is assumed to be
sufficiently lower than the signal, following the idea used in
\cite{Yarowsky92}.

The collected statistics were bigram and trigram occurrences.

\medskip

The algorithms and models were tested against a hand-disambiguated
menchmark corpus of over 6,500 words.

\medskip

We measure the performance of the different models in terms of recall
and precision.  Recall is the percentage of words that get the correct
tag among the tags proposed by the system. Precision is the percentage
of tags proposed by the system that are correct.

\begin{table}[ht] \centering
\begin{tabular}{|l|l|l|} \hline
           &{\bf CG-2 parser}          &{\bf Rel. Labelling} \\
           &{\bf prec. - recall} &{\bf prec. - recall} \\ \hline
   {\bf C} &\(90.8\%-99.7\%\)    &\(93.3\%-98.4\%\)    \\ \hline
\end{tabular}
\caption{Results obtained with the linguistic model.}
\label{table-results-L}
\end{table}

\begin{table}[ht] \centering
\begin{tabular}{|l|l|} \hline
            &{\bf Rel. Labelling}  \\
            &{\bf prec. - recall}  \\ \hline
   {\bf B}  &\(87.4\%-88.0\%\)     \\ \hline
   {\bf T}  &\(87.6\%-88.4\%\)     \\ \hline
   {\bf BT} &\(88.1\%-88.8\%\)     \\ \hline
\end{tabular}
\caption{Results obtained with statistical models.}
\label{table-results-S}
\end{table}

\begin{table}[ht] \centering
\begin{tabular}{|l|l|} \hline
             &{\bf Rel. Labelling} \\
             &{\bf prec. - recall} \\ \hline
  {\bf BC}   &\(96.0\%-97.0\%\)    \\ \hline
  {\bf TC}   &\(95.9\%-97.0\%\)    \\ \hline
  {\bf BTC}  &\(96.1\%-97.2\%\)    \\ \hline
\end{tabular}
\caption{Results obtained with hybrid models.}
\label{table-results-H}
\end{table}

Precision and recall results (computed on all words except puntuation
marks, which are unambiguous) are given in tables
\ref{table-results-L}, \ref{table-results-S} and
\ref{table-results-H}. Models are coded as follows: {\bf B} stands for
bigrams, {\bf T} for trigrams and {\bf C} for hand-written
constraints. All combinations of information types are tested. Since
the CG-2 parser handles only Constraint Grammars, we cannot test this
algorithm with statistical models.

These results suggest the following conclusions: 

\nobreak
\begin{itemize}
\item Using the same language model (107 rules), the relaxation
  algorithm disambiguates more than the CG-2 parser.  This is due to
  the weighted rule application, and results in more misanalyses and
  less remaining ambiguity.
\item The statistical models are clearly worse than the linguistic
  one. This could be due to the noise in the training corpus, but it
  is more likely caused by the difficulty of the task: we are dealing
  here with shallow syntactic parsing, which is probably more
  difficult to capture in a statistical model than e.g.\ POS tagging.
\item The hybrid models produce less ambiguous results than the other
  models. The number of errors is much lower than was the case with
  the statistical models, and somewhat higher than was the case with
  the linguistic model. The gain in precision seems to be enough to
  compensate for the loss in recall\footnote{This obviously depends on
    the flexibility of one's requirements.}.
\item There does not seem to be much difference between BC and TC
  hybrid models. The reason is probably that the job is mainly done by
  the linguistic part of the model -- which has a higher relative
  weight -- and that the statistical part only helps to disambiguate
  cases where the linguistic model doesn't make a prediction.  The BTC
  hybrid model is slightly better than the other two.
\item The small difference between the hybrid models suggest
  that some reasonable statistics provide enough disambiguation, and
  that not very sophisticated information is needed.
\end{itemize}

\section{Discussion}

In this paper we have presented a method for combining linguistic
hand-crafted rules with statistical information, and we applied it to
a shallow parsing task.

Results show that adding statistical information results in an
increase in the disambiguation ratio, getting a higher precision. The
price is a decrease in recall. Nevertheless, the risk can be
controlled since more or less statistical information can be used
depending on the precision/recall tradeoff one wants to achieve.

\medskip

We also used this technique to build a shallow parser with minimal
human effort:

\begin{itemize}
  
\item 107 disambiguation rules were written in a day.
    
\item These rules were used to analyze a training corpus, with a very
  high recall and a reasonable precision.

\item This slightly ambiguous training corpus is used for collecting
  bigram and trigram occurrences. The noise introduced by the
  remaining ambiguity is assumed not to distort the resulting
  statistics too much.

\item The hand-written constraints and the statistics are combined
  using a relaxation algorithm to analyze the test corpus, rising the
  precision to $96.1\%$ and lowering the recall only to $97.2\%$.

\end{itemize}

\noindent Finally, a reservation must be made: what we have not investigated in
this paper is how much of the extra work done with the statistical
module could have been done equally well or even better by spending
e.g.\ another day writing a further collection of heuristic rules. As
suggested e.g.\ by Tapanainen and Voutilainen (1994) and Chanod and
Tapanainen (1995), hand-coded heuristics may be a worthwhile addition
to `strictly' grammar-based rules.

\section*{Acknowledgements}

The authors wish to thank Pasi Tapanainen and two ANLP'97 referees for
useful comments.

The first author benefited from the collaboration of Juha Heikkil\"a
in the development of the linguistic description used by the EngCG
morphological tagger; the two-level compiler for morphological
analysis in EngCG was written by Kimmo Koskenniemi; the recent version
of the Constraint Grammar parser (CG-2) was written by Pasi
Tapanainen. The Constraint Grammar framework was originally proposed
by Fred Karlsson.

\end{document}